\documentclass[letterpaper, 10 pt, conference]{ieeeconf}  
  \usepackage{pgfplots}
  \pgfplotsset{compat=newest}
  \usetikzlibrary{plotmarks}
  \usetikzlibrary{arrows.meta}
  \usepgfplotslibrary{patchplots}
  \usepackage{grffile}
  \usepackage{amsmath}
  \usepackage{blindtext}
  \usepackage[bottom]{footmisc}
  \usepackage{mathrsfs}
  \usepackage{lipsum}
 \usepackage{url}
 \usepackage{array}
\usepackage{booktabs}
\usepackage{float}
\usepackage{multirow}
\usepackage{cite}

\usepackage{svg}

\newlength\fwidth
 \usepackage[linesnumbered,ruled]{algorithm2e}

\usepackage{graphicx}
\usepackage{framed}
\usepackage{indentfirst}
\usepackage{multirow}
\usepackage{makecell}

\usepackage{amsmath, amssymb,amsthm}
\usepackage{color}
\usepackage{booktabs}
\usepackage{tabularx}
\usepackage{hyperref}
\usepackage[capitalise]{cleveref}
\usepackage{comment}

\usepackage{makecell}
\let\labelindent\relax
\usepackage{enumitem}

\setcellgapes{1pt}

\theoremstyle{exampstyle}
\newtheorem{hypothesis}{H}
\usepackage{hyperref}

\newcommand{\eref}[1]{Eq.~\eqref{#1}}  
\newcommand{\sref}[1]{Sec.~\ref{#1}}    
\newcommand{\figref}[1]{Fig.~\ref{#1}}  
\newcommand{\prg}[1]{\noindent\textbf{#1}} 

\IEEEoverridecommandlockouts                              

\usepackage[colorinlistoftodos]{todonotes}

\usepackage[font=footnotesize]{caption}

\title{\LARGE \bf
Learning User-Preferred Mappings for Intuitive Robot Control
}

\author{Mengxi Li, Dylan P. Losey, Jeannette Bohg, and Dorsa Sadigh
\thanks{The authors are affiliated with the Computer Science Department, Stanford University, Stanford, CA 94305. \newline
{e-mail: \{mengxili, dlosey, bohg, dorsa\}@stanford.edu}}}

\begin{document}

\maketitle
\thispagestyle{empty}
\pagestyle{empty}

\begin{abstract}
When humans control drones, cars, and robots, we often have some preconceived notion of how our inputs should make the system behave. Existing approaches to teleoperation typically assume a \textit{one-size-fits-all} approach, where the designers pre-define a mapping between human inputs and robot actions, and every user must adapt to this mapping over repeated interactions. Instead, we propose a \textit{personalized} method for learning the human's preferred or preconceived mapping from a few robot queries. Given a robot controller, we identify an alignment model that transforms the human's inputs so that the controller's output matches their expectations. We make this approach data-efficient by recognizing that human mappings have strong \textit{priors}: we expect the input space to be proportional, reversable, and consistent. Incorporating these priors ensures that the robot learns an intuitive mapping from few examples. We test our learning approach in robot manipulation tasks inspired by \textit{assistive} settings, where each user has different personal preferences and physical capabilities for teleoperating the robot arm. Our simulated and experimental results suggest that learning the mapping between inputs and robot actions improves objective and subjective performance when compared to manually defined alignments or learned alignments without intuitive priors. The supplementary video showing these user studies can be found at: \textcolor{orange}{\url{https://youtu.be/rKHka0_48-Q}}
\end{abstract}

\section{Introduction}
\label{intro}
Humans are extremely good at developing tools and systems. Each of us interact with these systems everyday---from driving your car to work to playing video games at home. Unfortunately, we are not always good at these interactions. Using a joystick to play a kart racing computer game seems relatively easy, and we can do this without thinking much about it. But actually learning how to drive a real-life car does require practice. Even more challenging is flying a helicopter---this is highly demanding, and requires professional training. In these systems the human must adapt to the robot's controller across multiple rounds of interaction. Rather than forcing the human to adapt to the robot, we wonder if instead the robot could intelligently adapt to our preferences, making control more easy and intuitive? 

Alongside rapid advancements in the field of robotics, intuitive control is increasingly in demand, particularly since a wide spectrum of users are operating robots beyond just engineers~\cite{arata2018intuitive}. User-friendliness and comfort are crucial in many application domains including surgical and assistive robots~\cite{argall2018autonomy, robinson2016adaptable}, mobile robots~\cite{melidis2018intuitive} and even more abstract smart home systems~\cite{huang2017over}. At the heart of these systems, there is seamless \textit{teleoperation}. Typically---for teleoperation---there are two main categories of control methods. One is to use control interfaces such as joysticks or sip-and-puff devices \cite{boboc2012review, javdani2018shared}. These controllers are light-weight but low-dimensional, which makes control of robots with many {\em degrees of freedom} (DoF) challenging. To make up for this limitation when teleoperating robot arms, \cite{herlant2016assistive} propose to that the controller should change modes between different DoFs. Other work focuses on capturing high-DoF human body movement with wearable devices, cameras, or sensors~\cite{rebelo2014bilateral}, and then maps them to robot actions. While these methods provide more accurate measurements and natural control interfaces, they are typically larger and more expensive systems that might not be available to everyday users.

\begin{figure}[t]
    \vspace{0.5em}
	\begin{center}
		\includegraphics[width=1.0\columnwidth]{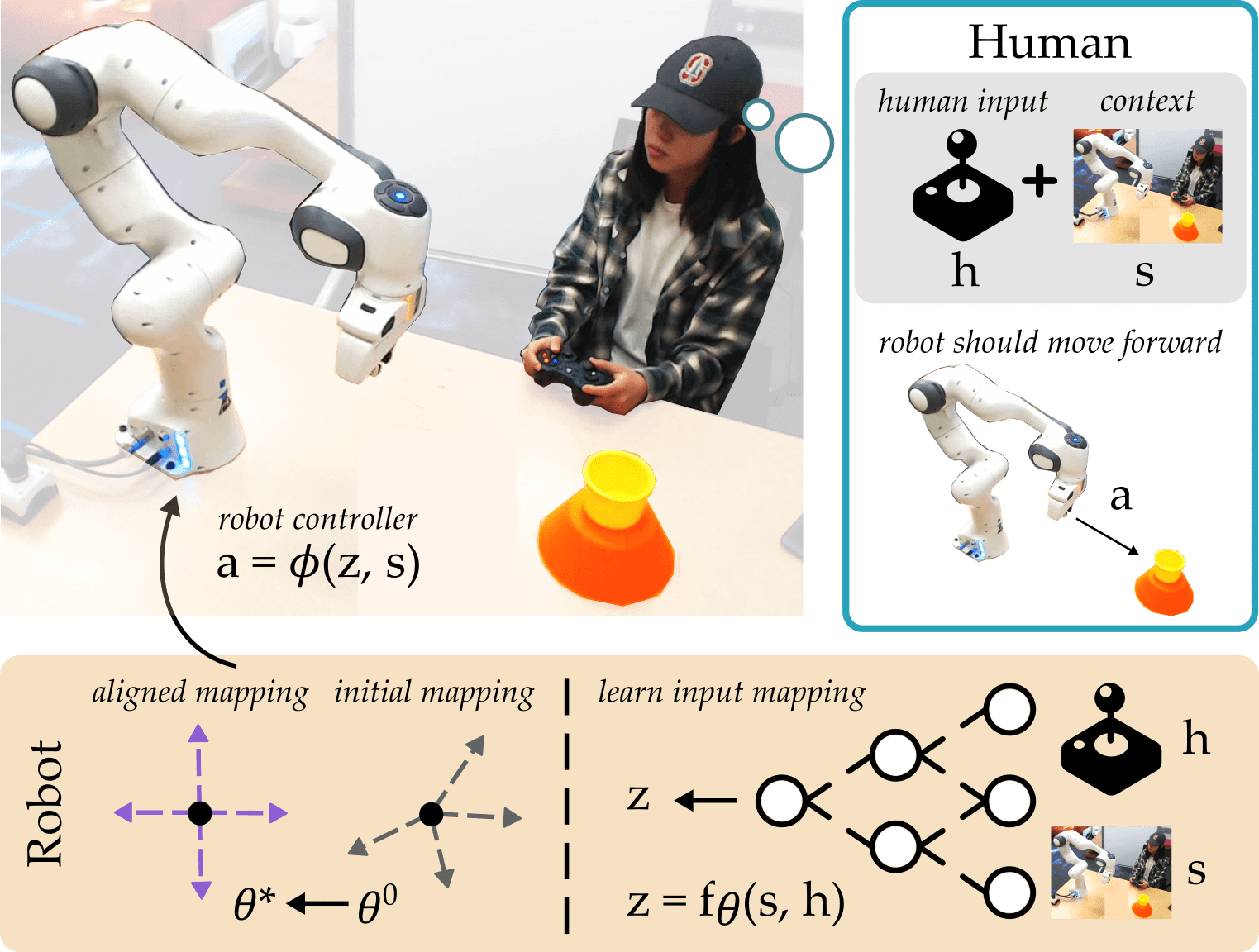}

		\caption{The human has in mind a preferred mapping between their joystick inputs and the robot's actions. The robot learns this mapping (i.e., $\theta$) offline from labelled data and intuitive priors, so that the robot's online actions are correctly aligned with the human's intentions.}

		\label{fig:front}
	\end{center}
	
	\vspace{-2em}
	
\end{figure}

Within this work, we are interested in the first category of teleoperation methods. We present a human-centered, adaptive system for robotic manipulation so that human users can smoothly and intuitively teleoperate high-DoF robots with simple, low-DoF controllers. We envision a model that provides a general framework for learning intuitive input mappings that are agnostic to both the controller and the dynamics of the underlying robotic system (see Fig.~\ref{fig:front}). Based on a general architecture of feedforward neural networks, the presented framework can be applied to different robots and controllers for various tasks with only a few demonstrations needed from the human users. Our key insight is: 

\vspace{-0.5em}
\begin{center}
\emph{By incorporating the intuitive priors that humans expect, including proportionality, reversability and consistency, we can quickly learn how humans prefer to control robots.} 
\end{center}\vspace{-0.5em}

Our approach first asks the human user to label a set of robot actions with their preferred inputs. For example, the robot moves towards the cup, and then asks what joystick direction should be associated with this action. Based on a small number of these labeled state-action pairs, we learn an alignment model that maps from human inputs to robot inputs such that the resulting robot actions match the human's preferences. Ensuring a \textit{small number of questions} is critical for real-world implementation: we cannot ask the human to label hundreds of examples actions! Our insight---incorporating intuitive priors---enables the robot to generalize across the workspace, leading to offline learning from a limited amount of labeled data.

More precisely, we make the following contributions:

\prg{Formalizing Priors for Human Control.} 
We formalize the properties that humans expect to have when  controlling robotic systems, including proportionality, reversability, and consistency. These priors are inspired by~\cite{jonschkowski2014state}, and we here analyse their relative importance when learning the human's preferred input mapping.

\prg{Generalizable Data-Efficient Learning.}
We build a general framework for learning human control preferences. We develop this data-efficient method by incorporating the intuitive priors that humans expect, so that users only need to provide a few examples of their desired mappings. Our proposed method develops a light-weight solution that can generalize across different controllers, dynamics, and tasks.  

\prg{Evaluating Learned Human Alignment.}
We implement our approach both in simulation and on robot manipulation tasks, and get subjective feedback from users. Our tasks are inspired by assisitve robotics, where users living with physical disabilities have different capabilities and preferences. Results in simulation as well as in user studies suggest that our algorithm successfully learned the human-preferred alignment between the low-DoF control interface and high-DoF robot actions. In practice, this learned alignment led to improved control and more seamless interaction.
\section{Related Work}
\label{sec:related_work}

\noindent \textbf{Intuitive Control.}
When controllers are intuitive, the effects of user inputs match with that user's expectations. Such a property is very important for robotic control, especially in human-robot teleoperation. Many research efforts are devoted to tackle this problem~\cite{duchaine2009safe, guenard2005dynamic,ciocarlie2009hand}. Some use wearable devices to capture human motion patterns~\cite{ribeiro2018uav}, some leverage augmented reality and use a gesture-based system to enable intuitive human robot teaming~\cite{gregory2019enabling}, and others rely on various sensors to measure and understand humans' haptic feedback~\cite{hussain2015vibrotactile}. These previous works require additional devices to infer the human intent behind their actions, and then provide a universal solution for intuitive control. Instead of directly enforcing a \textit{one-size-fits-all} control strategy to human users, we separately query each user for their \textit{individual} preferences. With no additional hardware requirements, we come up with a light-weight solution for intuitive control that could generalize across different platforms and tasks. \smallskip

\noindent \textbf{Assistive Robots.}
Intelligent assistive robot systems (e.g., wheelchair-mounted robot arms) are one important setting in which intuitive control is essential for widespread use ~\cite{yamazaki2010recognition}. Because users are constrained by person-specific capabilities, one-size-fits-all controllers are insufficient. Moreover, the problem of ``dimensionality mismatch'' arises when humans try to teleoperate these high-DoF robots by manipulating a low-DoF controller, e.g., a 2-axis joystick. Prior works tackle this with a mode switching mechanism \cite{nuttin2002selection, tsui2008development}: however, this forces users to constantly switch modes when performing complex tasks. 
Other works mitigate mode-switching with model-based optimization~\cite{herlant2016assistive} or in a data-driven manner~\cite{jain2016robot}, but the underlying mode switching still remains discontinuous. An alternative was recently proposed by~\cite{losey2019controlling}, in which the robot learns a continuous mapping between a low-DoF latent action space and the high-DoF robot action space using autoencoders. While this approach enables continuous control of the high-DoF robot, it still fails to provide an intuitive control mapping to the user. 

In this paper, we use \textit{assistive robot manipulation} as the test domain for our framework. We consider the teleoperation problem from the user's perspective, and attempt to learn a mapping between their inputs and the assistive robot's actions which caters to the user's preferences. 
\begin{figure*}[t]
	\begin{center}
		\includegraphics[width=2.0\columnwidth]{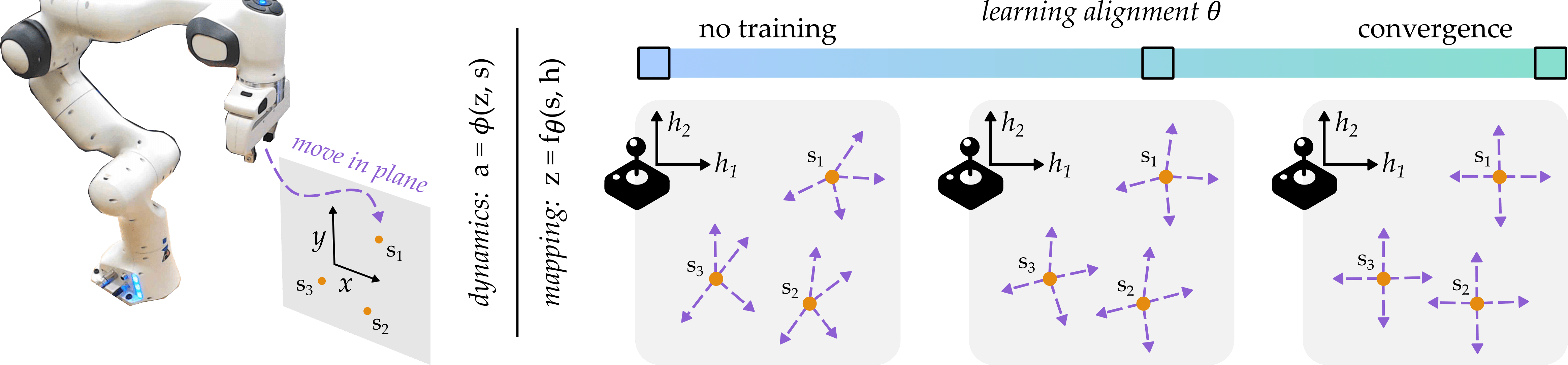}

		\caption{Visualization for the training process of the alignment model $z = f_\theta (h, s)$ parametrized by $\theta$. Here the example task is for the robot to move in a plane, and the human would like the robot's end-effector motion to align with their joystick inputs ($h_1$ and $h_2$). We take snapshots at three different points during training, and plot how the robot actually moves when the human presses up, down, left, and right. Note that this alignment is state dependent. As training progresses, the robot learns the alignment $\theta$, and the robot's motions are gradually and consistently pushed to match with the human's preferences.}

		\label{fig:align}
	\end{center}
	
	\vspace{-2em}
	
\end{figure*}

\section{Formalism for Modeling \\ Preference Alignment}
\label{sec:formalism}
We formulate a robot manipulation task as a discrete-time Markov Decision Process (MDP) $\mathcal{M}=\left(\mathcal{S}, \mathcal{A}, \mathcal{T}, R, \gamma, \rho_{0}\right)$. Here $\mathcal{S} \subseteq \mathbb{R}^{n}$ is the state space, $\mathcal{A} \subseteq \mathbb{R}^{m}$ is the robot action space, $\mathcal{T}(s, a)$ is the transition function, $R$ is the reward, $\gamma$ is the discount factor, and $\rho_{0}$ is the initial distribution.

\subsection{Problem Statement.} 
\label{sec:ps}
Assume humans control a robot using the teleoperation interface $a_t = \phi(z_t, s_t)$, where $t$ denotes timestep, $s_t \in \mathcal{S}$ is the system state, $z_t \in \mathbb{R}^d$ is the $d$-dimensional system input, and $a_t \in \mathcal{A}$ is the action executed by the robot. We use $h_t \in \mathbb{R}^d$ to denote the \textit{human's input}. For instance, when the human is using a joystick to teleoperate the robot arm, their two-DoF joystick input corresponds to $h \in \mathbb{R}^2$.

Traditionally, the human's input $h$ is directly used as the system input $z$, so that $z_t=h_t$, and the robot takes action $a_t = \phi(h_t, s_t)$. However --- for complex or non-intuitive systems --- it may be hard for human users to directly interact with the controller $\phi$. For instance, using a low-DoF joystick to control a high-DoF assistive robot can be difficult, since we do not know how to coordinate the robot's joints. Moreover, different users also have different preferences for controlling their robot---which may or may not match with the system controller. Put another way, the pre-defined mapping of the controller $a_t = \phi(h_t, s_t)$ is often quite different from what users want!

Our goal is to make the robotic system easier to control: instead of forcing the human to adapt to $\phi$, we want the robot to adapt to the user's preferences (without fundamentally changing the controller). We therefore propose to learn an \textit{alignment function} $z_t = f_\theta(h_t, s_t)$ parameterized by $\theta$, which maps the human input $h_t \in \mathbb{R}^d$ and the robot state $s_t \in \mathcal{S}$ to the controller input $z_t \in \mathbb{R}^d$. In this way, we construct a new, two-step mapping between the action of the robotic system $a_t$ and the human's input $h_t$:
\begin{equation}
    a_t =  \phi(z_t, s_t) = \phi(f_\theta(h_t, s_t), s_t)
\label{eqn:controller}
\end{equation}

\noindent \textbf{State Conditioning.} Consider the person in Fig.~\ref{fig:front}, who is using a 2-axis joystick to control a high-DoF assistive robot arm to reach and pour a cup. The user's preferred way to control the robot is unclear: what does the user mean if they push the joystick forward? When the robot is far from the cup, the user might intend to move the robot towards the cup---but when the robot is holding the cup, pressing forward now indicates that the robot should rotate, and pour the cup into a bowl! This mapping from the user input to intended action is not only person dependent, but it is also \textit{state dependent}. In practice, this state dependency prevents us from learning a single transformation to uniformly apply across the robot's workspace---we need an intelligent strategy for understanding the human's preferences in different contexts.

\subsection{Background: Latent Action Embeddings}

We will leverage the assistive robot controller proposed in \cite{losey2019controlling} as the main test domain for our alignment framework. Here a conditional autoencoder is trained from demonstrations of related tasks, learning the control function $\phi$ in \eref{eqn:controller}. More formally, $\phi : \mathcal{Z} \times \mathcal{S} \mapsto \mathcal{A}$ is a decoder that recovers a high-DoF robot action $a_t \in \mathcal{A}$ given the system input $z_t \in \mathcal{Z}$ and current context $s_t \in \mathcal{S}$. Overall, $\phi$ is a suitable test domain for our work since it is not immediately clear what the system inputs map to, i.e., there is no prior over how $z_t$ is aligned with $a_t$ at different states. We also point out that this controller captures the state conditioning described above. Since $\phi$ depends on $s$, the robot's action $a$ changes based on the current context: the same input $z$ moves the robot towards the cup when the gripper is empty, and then rotates to pour when holding the cup. This enables easy switching between tasks, like reaching and pouring.
\section{Approach}
\label{sec:approach}
\subsection{Constructing Alignment Model}
\label{sec:approach:align}
\prg{Human Alignment.} As described in \sref{sec:formalism}, we seek to obtain a mapping $z_t = f_{\theta}(h_t, s_t)$ that converts the human input $h_t$ to the controller input $z_t$ at timestep $t$. Using this transformed system input, the controller will then execute action $a_t$ on the robot, as in \eref{eqn:controller}. \smallskip

\noindent \textbf{Model.} We leverage a function $f_{\theta}(h_t, s_t)$ with parameters $\theta$ to capture this alignment. Here $s_t$ --- the current state of the robot --- is also passed as part of the input since the human's alignment could be state-dependent. In this work, we will utilize a general Multi-Layer Perceptron (MLP) to represent $f_\theta$, where $\theta$ becomes the weights of the network. \smallskip

\prg{Objective Function.}
Given the alignment model $f$, we will apply supervised learning to match the output of this model to the true human preferences (as visualized in Fig.~\ref{fig:align}). 

Formally, with the action $a_t$ computed by \eref{eqn:controller},  the robot state $s_{t+1}$ at next timestep is given by the transition model $s_{t+1} = \mathcal{T}(s_t, a_t)$. We therefore denote the overall mapping between the current human input $h_t$ at state $s_t$ and the next state $s_{t+1}$ using a function $T$ with parameters $\theta$:
\begin{equation}
    s_{t+1} = T_\theta(h_t, s_t) =  \mathcal{T}(s_t, \phi(f_{\theta}(h_t, s_t), s_t))
    \label{eqn:next_state_from_h}
\end{equation}
Because we are ultimately interested in how well the robot's action $a_t$ matches the human's preference, we define our objective function as the distance $d$ between the robot's actual next state $s_{t+1}$ and the ground truth state $s_{t+1}^{*}$, which is where the human really \textit{intended} to go\footnote{We query the user for this intended state, as explained in Section \ref{sec:approach:data_generation}}. Taking the expectation, we get the supervised loss $L_{sup}$:
\begin{equation}
    \label{eqa:supervised_loss}
    L_{sup}(\theta) = \mathbb{E}[d(s_{t+1}^{*}, s_{t+1})] = \mathbb{E}[d(s_{t+1}^{*}, T_\theta(h_t, s_t))]
\end{equation}
We leverage Euclidean Distance as our distance metric between states, although other metrics are also possible: $d(s_1, s_2)=|| s_1- s_2 ||_2$. To approximate the integral in \eref{eqa:supervised_loss}, we implement Monte Carlo Sampling by stochastically picking $N$ data points from the state distribution:
\begin{equation}
    \label{eqa:mc_supervised_loss}
    L(\theta) = \frac{1}{N}\sum_{i=1}^{N} d(s_{t+1}^{*i}, T_\theta(h_t, s^i_t))
\end{equation}

\subsection{Ensuring Data-Efficiency with Intuitive Priors}
Recall that we are learning the human's preferred mapping with a function approximator. In general, training these models requires a large number of labeled data samples, particularly in complex scenarios where the mapping changes in different states. However, since we need annotations from humans to identify their intended states, it is impractical for the robot to collect a large labeled dataset. Accordingly---to tackle the challenge of insufficient labeled data---we employ a \textit{semi-supervised} learning method~\cite{chapelle2009semi}. Our contribution here is to formulate the intuitive priors that humans have over their control mappings as \textit{loss terms}, which can then be leveraged within this semi-supervised learning. 

Formally, for a given robotic arm, we let $s$ be the robot's joint position, and we denote the forward kinematics as $\Psi$. For a human control input $h$ at state $s$, the next state is given by $T_\theta(h, s)$ as in \eref{eqn:next_state_from_h}. The corresponding end-effector pose change is  $\Delta x(s) =  \Psi(T_\theta(h, s)) - \Psi(s)$. With these definitions in mind, we argue an intuitive control mechanism should satisfy the following properties: \smallskip

\begin{enumerate}[leftmargin=0.5cm]
    \item \textit{Proportionality} -- The amount of change in the 3D pose of the robot's end effector should be proportional to the scale of the human input, i.e.,
\begin{equation*}
        \alpha \cdot | \Psi(T_\theta(h, s)) - \Psi(s) | = | \Psi(T_\theta(\alpha \cdot h, s)) - \Psi(s) |
\end{equation*}
     where $\alpha \in \mathbb{R}$ is the scaling factor. Proportionality is quite common and intuitive in system design: it indicates that humans expect the system to be linearly interpolatable. We define the proportionality loss $L_{prop}$ as:
    \begin{equation*}
    L_{prop} = \big[ \Psi(T_\theta(\alpha \cdot h, s)) - (\Psi(s) + \alpha \Delta x(s))\big]^2
    \end{equation*}
    where $\alpha$ is independently sampled from a uniform distribution $\alpha \sim U(-1, 1)$, since the control input is bounded.\smallskip
    
    \item \textit{Reversability} -- If an action $h$ makes the robot transit from state $s_1$ to $s_2$, the opposite action, denoted by the negation of $h$, i.e., $-h$, should be able to make the robot transit from $s_2$ back to $s_1$.
    \begin{equation*}
        s_2 = T_\theta(h, s_1) \rightarrow s_1 = T_\theta(-h, s_2)
    \end{equation*}
    This property ensures recoverability when users mistakenly operate the system, so that people can undo their mistakes and return to the original state. We define the reversability loss as:
    \begin{equation*}
    L_{reverse} = \big[\Psi(s) - \Psi(T_\theta(-h, T_\theta(h, s)))\big]^2
    \end{equation*}
    where $\Psi(s)$ is the current 3D pose of the robot end effector, while $\Psi(T_\theta(-h, T_\theta(h, s)))$ is the 3D pose of the end effector after executing human input $h$ followed by executing the opposite input $-h$. \smallskip
    
    \item \textit{Consistency} -- The same action taken at nearby states should lead to similar amounts of change in the 3D pose of the robot's end effector, i.e.,
\begin{gather*}
    \Delta x_1 = | \Psi(T_\theta(h, s_1)) - \Psi(s_1) | \\ \Delta x_2 = | \Psi(T_\theta(h, s_2)) - \Psi(s_2) |\\
    \forall \epsilon > 0, \exists \delta > 0, s.t. |s_1 - s_2| < \delta \rightarrow |\Delta x_1 - \Delta x_2| < \epsilon.
\end{gather*}
    Consistency encourages the control to be smooth, so that the mapping does not discontinuously change alignment. We define the consistency loss as:
    \begin{equation*}
    L_{con} = \exp{(-\gamma|| s_1 - s_2 ||)} \cdot (\Delta x(s_1) - \Delta x(s_2))^2
    \end{equation*}
    We use the weight term $\exp{(-\gamma || s_1 - s_2 ||)}$ to gauge the similarity between states $s_1$ and $s_2$, where ${\gamma > 0}$ is a hyperparameter controlling the temperature. A large weight is assigned to the state pair $(s_1, s_2)$ if the difference between them is small. 
\end{enumerate}

During the training process, we minimize the supervised loss for \textit{labeled} data combined with the semi-supervised loss for \textit{unlabeled} data using intuitive priors: 
\begin{equation}
    \label{eqn:all_loss}
    L = L_{sup} + \lambda_1 L_{prop} + \lambda_2 L_{reverse} + \lambda_3 L_{con}
\end{equation}
Here $L_{sup}$ is supervised loss term as shown in  Eq.~\eqref{eqa:supervised_loss}, and $\lambda_1, \lambda_2, \lambda_3$ are constant coefficients. Importantly, incorporating these different loss terms --- which are inspired by human priors over controllable spaces \cite{jonschkowski2014state} --- enables the robot to generalize the labeled human data (which it performs supervised learning on) to unlabeled states (which it can now perform semi-supervised learning on)!

\subsection{Data Collection by Querying Users}
\label{sec:approach:data_generation}
To enable the training of our algorithm, we need to collect labeled data tuples $(s_t, h_t, s_{t+1}^{*})$ as well as unlabeled data tuples $(s_t, s_{t+1}^{*})$, where $s_t$ is the robot's current state, $h_t$ is the human's low-dimensional input, and $s_{t+1}^{*}$ is the intended next robot state corresponding to $h_t$. 
We emphasize that the robot never needs to detect the human's intended state --- rather, the human labels robot actions with their preferred inputs.
Our data collection procedure is as follows:
\begin{itemize}
    \item \textbf{Step $\mathbf{0}$}: Implement the controller $a = \phi(s, z)$. This step is optional depending on the type of the controller; we want to emphasize that the controller will remain fixed, and will not be altered by our alignment model.
    \item \textbf{Step 1}: For the task of interest, we sample a valid state $s_t$ from the state distribution and randomly sample a valid controller input $z_t$ at state $s_t$.
    \item \textbf{Step 2}: At the sampled state $s_t$, we apply system input $z_t$ to the controller $\phi$ in order to get the robot action $a_t=\phi(s_t, z_t)$.
    \item \textbf{Step 3}: We record the subsequent robot state $s_{t+1}$ after executing the action $a_t$ at the state $s_t$. 
    \item \textbf{Step 4}: Steps 1, 2, 3 are repeated for $N$ iterations to get the unlabeled dataset $\{(s^i_t,  s_{t+1}^{*i})_{i=1}^N\}$, consisting of $N$ unlabeled (state, next-state) tuples.
    \item \textbf{Step 5}: We randomly sample $(s_t, s_{t+1}^{*})$ from the unlabeled dataset collected in Step 4, and then we query the humans for the corresponding labels $h_t$. Here the user provides examples to the robot of what controls they would find intuitive to move from $s_t$ to $s_{t+1}^{*}$.
    \item \textbf{Step 6}: Step 5 is repeated for $K$ iterations to get the labeled dataset $\{(s^i_t, h^i_t, s_{t+1}^{*i})_{i=1}^K\}$, consisting of $K < N$ labeled samples.
\end{itemize}

For supervised learning approaches, typically $K$ needs to be fairly large so that the state space is well covered. Here we learn the model with only a small number of labeled samples, and then generalize to other unlabeled states by exploiting intuitive control properties. However, we do point out that the number of queries needed could vary according to the task complexity. In addition, for query selection, more intelligent methods such as \textit{active learning} could be integrated to further improve training. Within this work, we will employ a simple strategy of \textit{passively} sampling from the state distribution. Even with this simple approach, we are able to learn the user's preferred mappings from a limited number of queries in both our simulations and the robot manipulation tasks, as we will show in the following sections.

\begin{figure*}[t]
	\begin{center}
		\includegraphics[width=2.0\columnwidth]{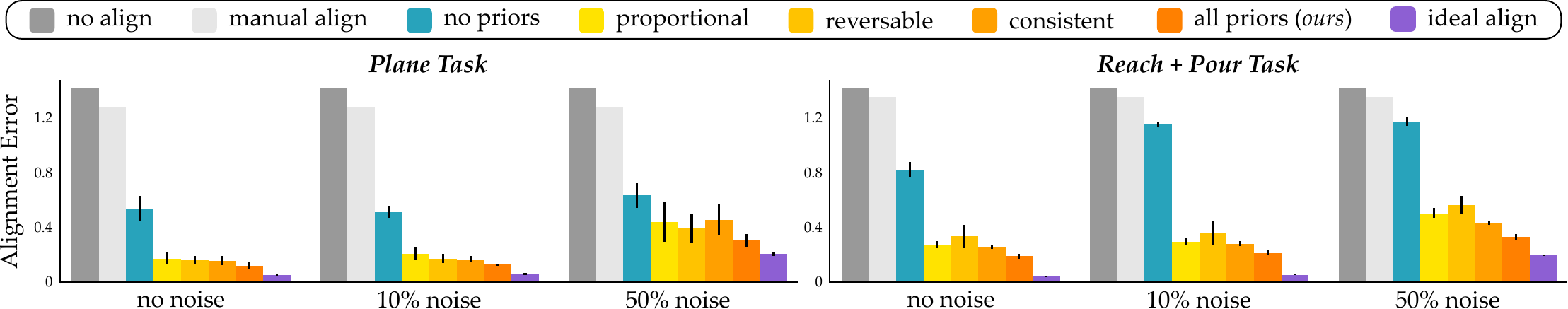}

		\caption{Quantitative results from the simulation experiments. We tested three different tasks with increasing complexity, and here we display the results of the easiest (\textit{Plane}) and the hardest (\textit{Reach + Pour}). \textit{Alignment Error} refers to a weighted sum of the relative positional and rotational error in end-effector space. To explore the robustness of our method, we additionally varied how noisy the human was when providing labels.  Across different tasks and levels of human noise, including all priors consistently outperformed the other methods, and almost matched an ideal alignment learned from abundant data.}
		\label{fig:sims}
	\end{center}
	
	\vspace{-2em}
	
\end{figure*}

\section{Simulation Experiments}
\label{sec:simulation_experiment}
To test our proposed alignment algorithm, we conducted simulations on a Franka Emika Panda robot arm for three manipulation tasks of increasing complexity. We will first discuss the details that are consistent across all experiments, and then elaborate on each task respectively. \smallskip 

\prg{Setup.} 
We used a simulated Panda robot arm. The state $s \in S$ denotes the robot’s joint position, and the action $a \in A$ refers to the robot’s joint velocity. The system transition function is $s^\prime = s + a\cdot dt$, where $dt$ is the step size. We trained a conditional autoencoder \cite{losey2019controlling} for each task by collecting trajectory demonstrations on the simulated robot. The learned decoder maps the low-dimensional system input $z \in \mathbb{R}^2$ to the high-dimensional robot action $a \in \mathbb{R}^7$ --- this decoder is used as the controller $\phi$ in our experiments. The alignment model $z = f_\theta(h, s)$ aims to identify the correspondence between control inputs $z$ and human inputs $h$ at each state $s$. Applying a simulated virtual human $H_{sim}$ that is separately defined for each task, we generated data for training and testing using the procedure described in Section~\ref{sec:approach:data_generation}. \smallskip
   
\begin{figure*}[t]
    \vspace{-0.5em}
	\begin{center}
		\includegraphics[width=2\columnwidth]{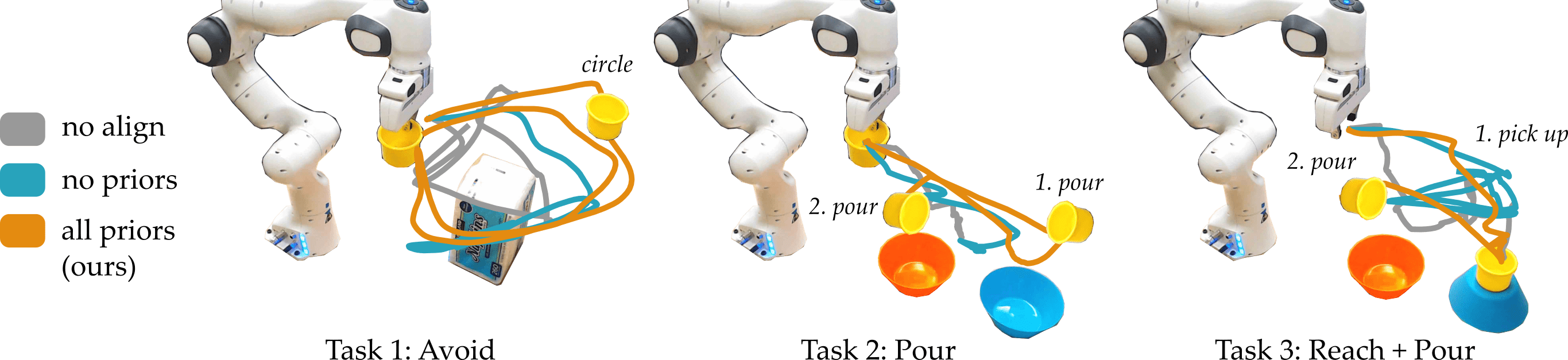}

		\caption{Example end-effector trajectories for the \textit{Avoid}, \textit{Pour} and \textit{Reach + Pour} tasks during our user studies. Participants teleoperated the 7-DoF Panda robot arm without any alignment model (\textit{no align}), with an alignment model trained only on their supervised feedback (\textit{no priors}), and with our proposed method, where the robot generalizes the human's feedback using intuitive priors (\textit{all priors}). For both baselines, we can see examples of the human getting confused, counteracting themselves, or failing to complete the task.}
		\label{fig:user_display_traj}
		
	\end{center}
	
	\vspace{-1.5em}

\end{figure*}

\prg{Tasks.} 
We considered three tasks with increasing levels of complexity; these tasks roughly correspond to our user study tasks shown in Fig.~\ref{fig:user_display_traj}. Across all tasks the simulated user was given a 2-DoF joystick, such that $h \in \mathbb{R}^2$. 
\begin{enumerate}[leftmargin=0.5cm]
\item \textit{Plane}: The simulated robot arm moves its end-effector in a 2-dimensional horizontal plane. In this task, the simulated human's preference is to use one dimension of the joystick for controlling the movement along the $x$ axis and the other dimension for controlling the movement along the $y$ axis (also see Fig.~\ref{fig:align}).
\item \textit{Pour}: The simulated robot arm moves and rotates its end-effector along the vertical axis. Here the simulated human's preference is to use one dimension of the joystick for controlling the position and to use the other dimension for controlling the rotation of the end-effector. 
\item \textit{Reach \& Pour}: The simulated robot arm reaches for a bottle and then lifts and rotates the bottle to pour it into a bowl. In this task, the simulated human's preference is divided into two parts, and is based on the previous tasks. For reaching the bottle in the 2D plane, the preference is defined as in the task \textit{Plane}. For pouring, the preference for joystick control is the same as in the task \textit{Pour}.
\end{enumerate}

\noindent\textbf{Data Efficiency.} The simulated human provides 10 labeled data samples $\{(s_t^i, h^i_t, s^{*i}_{t+1})_{i=1}^{10}\}$, and the robot collects 1000 unlabeled data samples $\{(s_t^j, s^{*j}_{t+1})_{j=1}^{1000}\}$ for the tasks \textit{Plane} and \textit{Pour}. For the  sequential task \textit{Reach \& Pour}, we have 20 labeled samples and 2000 unlabeled samples. \smallskip

\prg{Model Details.}
The teleoperation controller in our experiments follows the structure in \cite{losey2019controlling}, and the human alignment model is constructed as in Section~\ref{sec:approach:align}. Our human alignment model is a Multi-Layer Perceptron (MLP) network with 2 hidden layers. 
For supervised training, the loss function we adopt is given by $L = L_{trans} + \lambda L_{rot}$. Let $L_{trans}$ be the $L_2$ loss between predicted position and ground truth position, and let $L_{rot}$ be the $L_2$ loss defined on the quaternion representation between predicted and actual rotation. In the \textit{Plane} task, $\lambda = 0$ because rotation is involved: but for the other two tasks, $\lambda = 1$. During semi-supervised training, the loss function is a combination of supervised loss terms and semi-supervised loss terms as described in Eq.~\eqref{eqn:all_loss}. \smallskip

\prg{Independent Variables.}
Within each simulated task, we varied (a) the type of alignment model and (b) the noise level of the simulated human oracle $H_{sim}$. We compared against two baselines that did not learn an alignment: \textit{no align}, where $z = h$, and \textit{manual align}, where we applied an affine transformation that best matched the data across all states. To understand which priors are useful, we also performed an ablation study where the robot learned the mapping with one prior at a time. Finally, we included a \textit{no prior} baseline, which only leveraged supervised training. To test the model robustness to noisy human labels---where the human incorrectly matches $h$ to $a$---we set the coefficient of variance $\frac{\sigma}{\mu} \in\{0,0.1,0.5\}$ for the simulated human. \smallskip

\prg{Dependent Measures.}
In both our approach and baselines, we measured position and rotation errors. For positions, we computed the \textit{relative} distance error $E_d = \frac{\lVert  x_{t+1}^*-x_{t+1}\rVert}{\lVert  x_{t+1}^*-x_t\rVert}$, where $x_t$ is the current end-effector position before movement, $x_{t+1}^* = \Psi(s^*_{t+1})$ is the desired end-effector position corresponding to intended next state $s^*_{t+1}$, and $x_{t+1}$ is the actual end-effector position the robot reaches after executing action $a_t = \phi(s_t,  f_\theta(h_t, s_t))$. We also measured the distance between rotations $E_r=2 \arccos (|\langle q_{t+1}, q^*_{t+1}\rangle|)$, where $q_{t+1}$ and  $q^*_{t+1}$ are the quaternion representation of the predicted and ground truth orientation, respectively. For each experiment setting, we reported mean and standard deviation of the performance metrics over 10 total runs. \smallskip

\prg{Hypotheses.}
We have the following three hypotheses:
\begin{hypothesis}
    With abundant labeled data, the alignment model will accurately learn the human's preferences.
\end{hypothesis}
\begin{hypothesis}
    Compared to the fully-supervised baseline, our semi-supervised alignment models that leverage intuitive priors will achieve similar performance with far less human data.
\end{hypothesis}
\begin{hypothesis}
    Semi-supervised training with proportional, reversible, and consistent priors will outperform alignment models trained with only one of these priors.
\end{hypothesis}

\prg{Results \& Analysis.}
The simulations demonstrate that our alignment method successfully learned the human's preference with a limited amount of labeled data. As shown in \figref{fig:sims}, each of the proposed alignment models significantly outperformed the one-size-fits-all baselines. For models that do not leverage data or learning --- i.e., \textit{no align} and \textit{manual align} --- the error is significantly higher than learning alternatives. With abundant data and noise-free human annotations, \textit{ideal align} provided the best-case performance: indicating that our parametrization of the alignment model is capable of capturing the human's control preferences.

Of course, in practice the amount of human feedback is limited. We therefore focus on models that learned from only a small number of labeled datapoints ($10 - 20$ examples). Here our proposed priors were critical: semi-supervised models that included at least one intuitive prior performed twice as well as \textit{no priors}, the supervised baseline. 

Across different noise levels, our model that leverages \textit{all priors} consistently demonstrated the lowest mean error and standard deviation. This was particularly noticeable when the human oracle is noisy, suggesting that the three priors are indeed \textit{complementary}, and including each of them together brings a performance boost! 

Comparing the easier \textit{Plane} task to the more complex \textit {Reach \& Pour} task, we also saw that using priors became more important as the task got harder. This suggests that --- in complex scenarios --- simply relying on a few labeled examples may lead to severe overfitting. On the other hand, our intuitive priors could effectively mitigate this problem. \smallskip

\prg{Summary.} Viewed together, the results of our simulations strongly support the hypotheses \textbf{H1}, \textbf{H2}, and \textbf{H3}. Our proposed alignment model successfully learned the mapping between human actions and the system input space (\textbf{H1}). In settings with limited labels, our proposed alignment model with intuitive control priors reached results that almost match supervised training with abundant data (\textbf{H2}). Finally, in ablation studies, we showed how combining all three proposed priors leads to superior performance and greater training stability than training with a single prior (\textbf{H3}).

\section{User Studies}
\label{sec:user_study}

Within this section we show the results of a user study that evaluates our framework across three robot manipulation tasks based on assistive settings. Participants teleoperated a 7-DoF robotic arm  (Panda,
Franka Emika) through a handheld 2-axis joystick controller. As in our simulations, we use \cite{losey2019controlling} to learn a decoder that enables low-DoF control over the high-DoF robot arm. The robot learned the user's individual preferences for how this controller should behave by asking for a limited number of examples and then generalizing with our intuitive priors.

\prg{Tasks.} 
Similar to our simulation experiments, we considered three different tasks. These tasks are visualized in Fig.~\ref{fig:user_display_traj}.

\begin{enumerate}[leftmargin=0.5cm]
\item \textit{Avoid}: The Panda robot arm moves its end-effector within a 2-dimensional horizontal plane. Users are asked to guide the robot around an obstacle without colliding with it. The task ended once users completed one clockwise rotation followed by one counter-clockwise rotation. 

\item \textit{Pour}: The Panda robot arm is holding a cup, and users want to pour this cup into two bowls. Users are asked to first pour into the farther bowl, before moving the cup back to the start and pouring into the closer bowl.

\item \textit{Reach \& Pour}: This is the most complex task. Users start by guiding the robot towards a cup and then pick it up. Once the users reach and grasp the cup, they are asked to take the cup to a target bowl, and finally pour into it.

\end{enumerate}

\prg{Independent and Dependent Variables.} For each of the task described above, we compared three different alignment models: making no adjustments to the original controller (\textit{no align}), an alignment model trained just using the human's labeled data (\textit{no priors}), and an alignment model trained using our proposed semi-supervised approach (\textit{all priors}). Our semi-supervised learning model should generalize the human's preferences by enforcing intuitive priors such as proportionality, reversibility, and consistency.

To evaluate the effectiveness of these different alignment strategies, we recorded quantitative measures including task completion time and trajectory length. We also calculated the percentage of the time that users \textit{undo} their actions by significantly changing the joystick direction --- undoing suggests that the alignment is not quite right, and the human is still adapting to the robot's control strategy. Besides these objective measures, we also collected subjective feedback from the participants through 7-point Likert scale surveys. \smallskip

\prg{Hypothesis.} An alignment model learned from user-specific feedback and generalized through intuitive priors makes it easier for humans to control the robot and perform assistive manipulation tasks. \smallskip

\prg{Experimental Procedures. }
We recruited 10 volunteers that provided informed written consent (3 female, ages $23.7 \pm 1.5$). Participants used a 2-axis joystick to teleoperate the 7-DoF robot arm, and completed three manipulation tasks inspired by assistive settings. At the start of each task, we showed the user a set of robot movements and ask them to provide their preferred input on the joystick --- i.e., ``if you wanted the robot to perform the movement you just saw, what joystick input would you provide?'' Users answered $7$ queries for task \textit{Avoid}, $10$ queries for task \textit{Pour} and $30$ queries for task \textit{Reach \& Pour}. After the queries finished, the users started performing tasks sequentially using each of the alignment strategies. The order of alignment strategies was counterbalanced.\smallskip

\begin{figure}[t]
	\begin{center}
		\includegraphics[width=1.0\columnwidth]{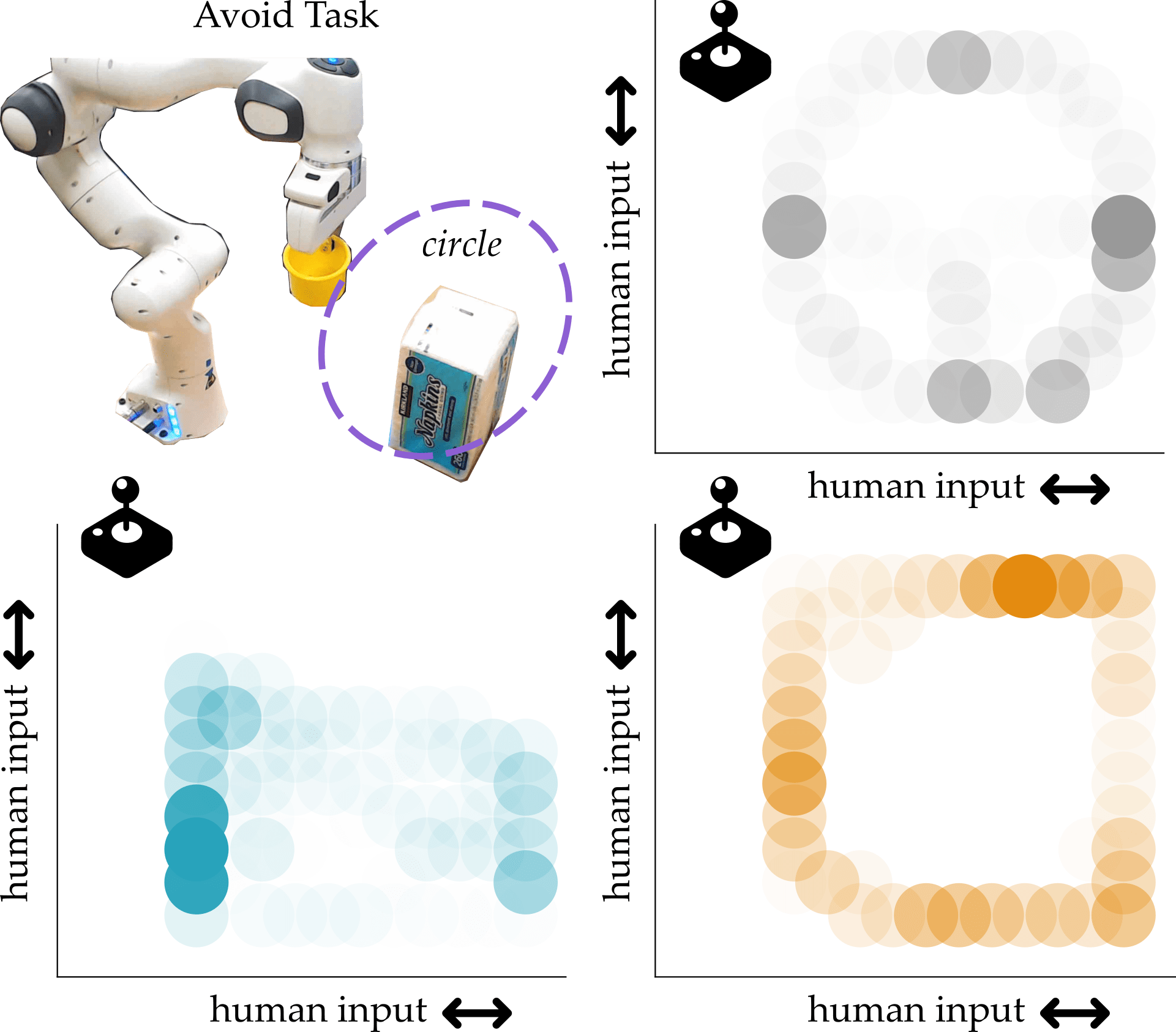}
		\caption{Heatmaps of the participants' joystick inputs during task \textit{Avoid}. For \textit{no align} in the upper right, people primarily used the cardinal directions. For \textit{no priors} in the bottom left, the joystick inputs were not clearly separated, and no clear pattern was established. 
		For our \textit{all priors} model on the bottom right, however, we observed that the human inputs were \textit{evenly distributed}. This indicates that the users smoothly completed the task by continuously manipulating the joystick in the range $[-1, +1]$ along both axes.}
		\label{fig:joystick}
	\end{center}
	\vspace{-2em}
\end{figure}

\begin{figure*}[t]
    \vspace{-0.5em}
	\begin{center}
		\includegraphics[width=2\columnwidth]{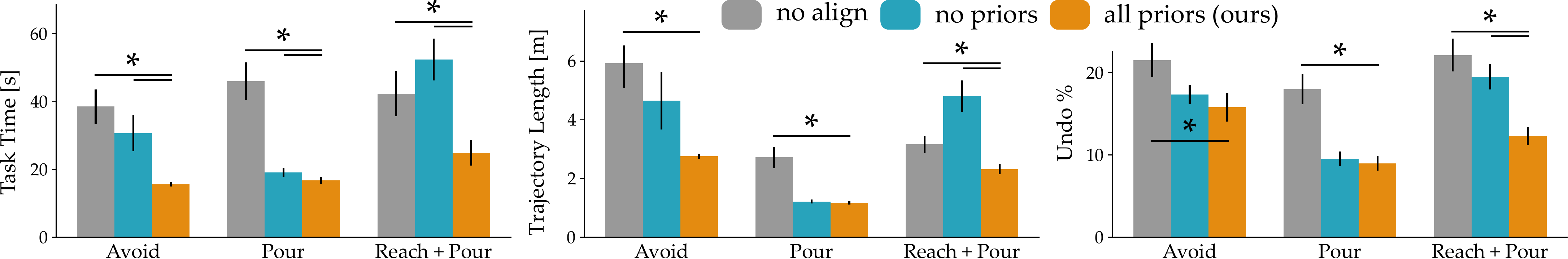}

		\caption{Objective results from our user study. Left: Average time taken to complete each task. Middle: Average trajectory length as measured in end-effector space. Right: Percentage of the time people spend undoing their actions. Error bars show the standard deviation across the $10$ participants, and colors match Fig.~\ref{fig:user_display_traj}. Asterisks denote statistically significant pairwise comparisons between the two marked strategies ($p < .05$).}
		\label{fig:objectives}
		
	\end{center}
	
	\vspace{-1em}

\end{figure*}

\prg{Results \& Analysis}
The objective results of our user study are summarized in Fig.~\ref{fig:objectives}. Across tasks and metrics, our model with \textit{all priors} outperforms the two baselines. In addition, our model not only has the best average performance, but it also demonstrates the least variance. Similar to our results in simulation, when the task is difficult (i.e., \textit{Reach \& Pour}), the performance of the align model trained with only supervised loss and \textit{no priors} drops significantly compared to simpler tasks, reinforcing the importance of priors in learning complex alignment models!

\begin{figure}[t]
    \vspace{-0.5em}
	\begin{center}
		\includegraphics[width=1\columnwidth]{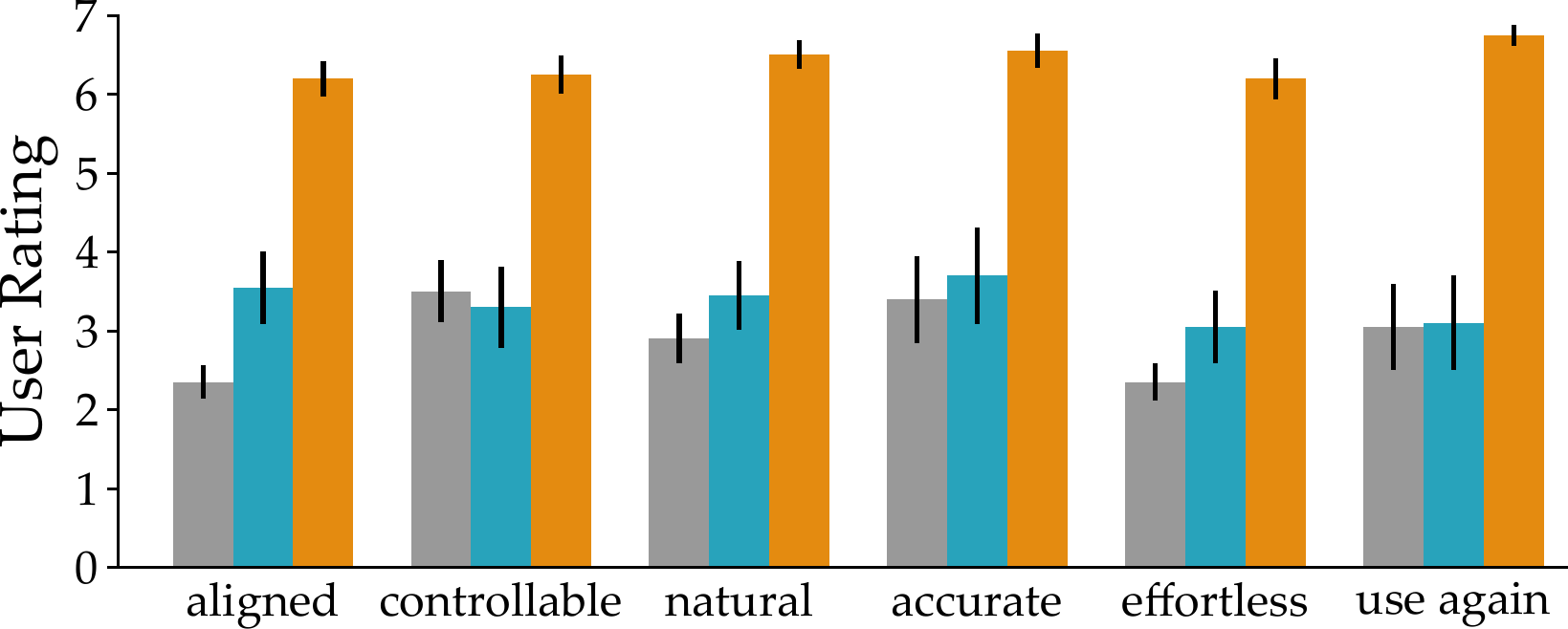}

		\caption{Results from our $7$-point Likert-scale survey. Higher ratings indicate agreement. Users thought that our learned model with intuitive priors aligned with their preferences, was easy to control, and improved efficiency --- plus they would choose to use it again! Pairwise comparisons between our approach and the baselines are statistically significant across the board.}
		\label{fig:user_subjective}
		
	\end{center}
	
	\vspace{-2em}

\end{figure}

We also illustrate our survey responses in Fig. \ref{fig:user_subjective}. Across the board, we found that users exhibited a clear preference for our proposed method. Specifically, they perceived \textit{all priors} as resulting in better alignment, more natural, accurate, and effortless control, and would elect to use it again. These subjective results highlight the importance of personalization when controlling high-DoF systems --- we contrast these results to \cite{losey2019controlling}, where participants perceived the unaligned controller as confusing and unintuitive.

To better visualize the user experiences, we also display example robot end-effector trajectories from one of the participants in Fig.~\ref{fig:user_display_traj}. Here we observe that the trajectories of our model (in orange) are smooth and do not detour during the tasks, while the trajectories for \textit{no align} (in grey) and \textit{no priors} (in blue) have many movements that counteract themselves, indicating that this user was struggling to understand and align with the control strategy. In the worst case, participants were unable to complete the task with the \textit{no priors} model (see the \textit{Avoid} task in Fig.~\ref{fig:user_display_traj}) because no joystick inputs mapped to their intended direction, effectively causing them to get stuck at undesirable states.

To further validate that our model is learning the human's preferences, we illustrate heatmaps over user inputs for the \textit{Avoid} task in \figref{fig:joystick}. Recall that this task requires moving the robot around an obstacle. Without the correct alignment, users default to the four cardinal directions (\textit{no align}), or warped circle-like motions (\textit{no priors}). By contrast, under \textit{all priors} the users smoothly moved the joystick around an even distribution, taking full advantage of the joystick's $[-1, +1]$ workspace along both axes. \smallskip

\prg{Summary.} Taken together, these objective measurements as well as subjective results empirically support our hypothesis. Using the alignment model learned with our intuitive priors, users are able to complete the manipulation tasks more efficiently, and in a way that matches their personal preferences.
\section{Discussion}
\label{discussion}

\prg{Summary.} We developed a framework for learning personalized mappings between human inputs and robot actions. Since no two humans are the same, we doubt that any one-size-fits-all approach will work well with everyday users. But asking humans about their preferences --- and getting their feedback in every situation --- is prohibitively time consuming. We therefore proposed a semi-supervised approach, where the robot has access to a few examples of the human's desired mapping, and must generalize that mapping across unlabeled data. We achieve this generalization by recognizing that humans have strong \textit{priors} over how controllers should operate: we expect the input space to be proportional, reversible, and consistent. By incorporating these priors, our proposed approach learned the human's preferences from a few queries. Importantly, this proposed method does not affect the system controller itself, is lightweight and agnostic to the underlying robot dynamics, and does not require any additional hardware or software for intent recognition.\smallskip

\prg{Limitations and Future Work.} The robot currently queries the human at random states. Future work will incorporate active learning, so that the robot intelligently selects informative states to ask for human preferences.





\section*{Acknowledgements}
We acknowledge funding from a Fanuc Fellowship, Qualcomm Innovation Fellowship, and the NSF Award \#1941722.

\bibliographystyle{IEEEtran}
\bibliography{references}

\end{document}